\documentclass[10pt,twocolumn,letterpaper]{article}
\usepackage[preprint]{cvpr}

\definecolor{cvprblue}{rgb}{0.21,0.49,0.74}
\usepackage[pagebackref,breaklinks,colorlinks,allcolors=cvprblue]{hyperref}
\def\confName{CVPR}
\def\confYear{2027}
\title{Can an Agent Trust Its Own Zoom?\\
Cross-Magnification Provenance Verification for Whole-Slide Images}
\author{
Yan Wang$^{1,*}$ \quad
Yiran Song$^{2,*}$ \quad
Zizhuo Gong$^{2}$ \quad
Jianxiang Qi$^{2}$\\
Zhongxing Xu$^{3}$ \quad
Lingfei Qian$^{1,\dagger}$ \quad
Mingquan Lin$^{2,\dagger}$\\[4pt]
$^{1}$Yale University, New Haven, CT, USA \quad
$^{2}$University of Minnesota -- Twin Cities, Minneapolis, MN, USA\\
$^{3}$Monash University, Melbourne, VIC, Australia\\[2pt]
{\small $^{*}$Equal contribution \quad $^{\dagger}$Corresponding authors:}
{\tt\small lingfei.qian@yale.edu}, {\tt\small lin01231@umn.edu}
}
\begin{document}
\maketitle
\begin{abstract}
Visual agents navigate whole-slide images through magnified observations associated with claimed locations. However, a visually plausible observation does not necessarily originate from its claimed location, particularly across magnifications and in regions with similar tissue structures. We formulate \textbf{cross-magnification provenance verification (CMPV)} to evaluate whether visual evidence supports a claimed spatial association, recover the source when possible, and abstain when evidence is insufficient. We introduce \textbf{PyraFault}, a benchmark of 3.06 million coordinate-supervised examples from 2,200 cases across five TCGA cohorts, covering within-context substitutions, absent-source faults, and geometric perturbations. We further propose \textbf{ZoomTrace}, a claim-independent verifier that estimates the observation's source before evaluating its agreement with the claimed location. It combines physical-scale alignment, visual retrieval, geometric refinement, and reliability estimation using spatially disjoint competing matches. On 503,320 test examples, ZoomTrace achieves 0.969 AUPRC and 0.987 AUROC, compared with 0.789 and 0.903 for Efficient LoFTR. Further evaluations reveal trade-offs in reliability estimation and limitations in vision-language models, cross-dataset retrieval, and calibration. Integrating ZoomTrace into visual agent tool paths reduces exposure to corrupted observations and silent failures, while downstream accuracy improvements depend on the agent and verification threshold. Our code and data are released at GitHub\footnote{\url{https://github.com/Yan2266336/PyraFault}}.
\end{abstract}

\section{Introduction}
\label{sec:intro}
\begin{figure*}[t]
    \centering
    \includegraphics[width=\textwidth]{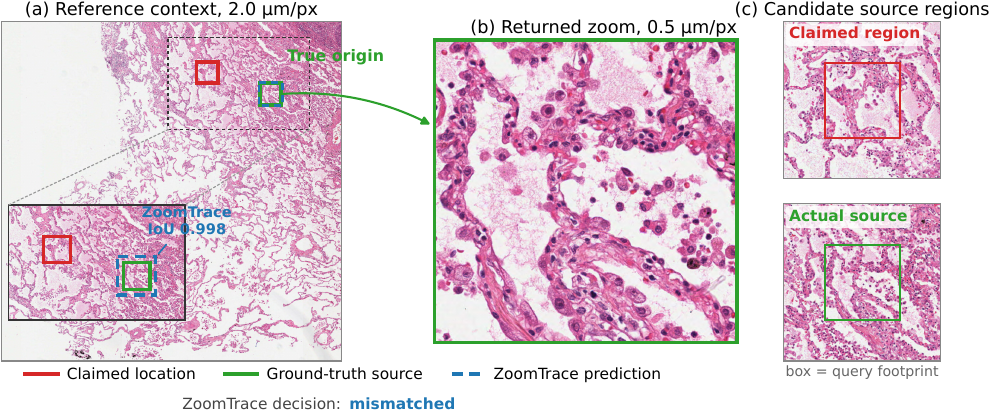}
    \caption{\textbf{Cross-magnification provenance verification
    in a whole-slide pathology image.}
    (a) The reference context shows the claimed location (red),
    ground-truth source (green), and ZoomTrace prediction
    (blue dashed).
    (b) The returned high-magnification observation originates
    from the true source rather than the claimed location.
    (c) The claimed and actual source regions exhibit visually
    similar tissue structures.
    ZoomTrace independently recovers the source
    (IoU = 0.998) and correctly identifies the provenance mismatch.}
    \label{fig:motivation}
\end{figure*}

Visual agents explore whole-slide images by requesting magnified views~\cite{chen2025pathagent,buckley2025giant,yang2026pathnavigate}. Each observation carries a claimed location that connects local evidence to its reference context. 
An agent may correctly interpret a magnified observation
yet attribute it to the wrong location, turning valid visual content into misleading spatial evidence. This raises a question: \emph{\textbf{when does a visual match provide sufficient evidence of a claimed spatial origin?}} Figure~\ref{fig:motivation} shows a plausible observation from another location with similar tissue structures. Such mismatches may arise from pyramid-coordinate conversion errors or incorrect content--metadata associations~\cite{hsu2025mainecoon,fraggetta2018importance}.

Matching and localization~\cite{sun2021loftr,wang2024eloftr} face three challenges here. \textbf{Magnification changes visual detail}: a $512\times512$ query occupies only $128\times128$ context pixels in our setting, where cellular details may be absent. \textbf{Repeated tissue structures support competing locations}, connecting provenance verification to ambiguity in visual place recognition~\cite{zaffar2024sue,miller2025doubt}. \textbf{Sources may lie outside the context}, elsewhere on the same slide or on another slide, even when an in-context match appears plausible. Therefore, verification must assess spatial agreement and permit abstention when evidence is insufficient.

We formulate \emph{cross-magnification provenance verification (CMPV)}: given a magnified query, lower-magnification context, and claimed location, evaluate the spatial claim, recover the source within the context when possible, and abstain under uncertainty. \emph{PyraFault} provides 3.06 million examples from 2,200 cases across five TCGA cohorts. Pyramid coordinates and slide identities establish true sources independently of claims. PyraFault
includes controlled provenance faults, appearance-mined alternatives, and geometric viewing perturbations. The test set samples all eight conditions equally, with 12.5\% correct claims.

Our verifier, \emph{ZoomTrace}, estimates the source and its reliability before consulting the claim. It aligns physical scale, retrieves candidates using frozen features, and refines locations through geometric matching. Building on matching uncertainty~\cite{zaffar2024sue,miller2025doubt}, reliability compares the strongest match with spatially disjoint alternatives; overlapping windows may represent the same source. The final decision combines source--claim agreement with reliability, enabling verification, mismatch detection, or abstention, and local source recovery for incorrect claims.

On 503,320 test examples, ZoomTrace reaches AUPRC 0.969 and AUROC 0.987, versus 0.789 and 0.903 for Efficient LoFTR. Ablations support scale alignment, refinement, and candidate diversity. Winning-score confidence gives stronger aggregate discrimination, while disjoint competition yields more uniform coverage across perturbations. Seven VLMs show limited claim-free localization in the tested settings; CAMELYON16 exposes retrieval and calibration limitations. Verification reduces exposure and silent failures; verdict gains vary by agent and threshold.

Our contributions are: (1) \textbf{CMPV}, a task for spatial provenance discrimination, local source recovery, and selective decisions across magnifications; (2) \textbf{PyraFault}, a coordinate-supervised benchmark of controlled provenance faults; and (3) \textbf{ZoomTrace}, a claim-independent verifier combining geometric localization with competing-source reliability, evaluated through component studies and visual-agent interventions.

\section{Related Work}
\label{sec:related}

\paragraph{Visual agents and evidence acquisition.}
V*~\cite{wu2024vstar} and ZoomEye~\cite{shen2025zoomeye}
guide detailed image exploration.
PathAgent~\cite{chen2025pathagent},
GIANT~\cite{buckley2025giant}, and
PathNavigate~\cite{yang2026pathnavigate}
extend navigation and evidence selection to gigapixel
pathology, while PathAgentBench~\cite{liao2026pathagentbench}
evaluates retrieval, localization, and multi-scale reasoning.
MedVIGIL studies recognition of broken visual evidence in
medical question answering~\cite{jiang2026medvigil}.
PathTrace audits recorded navigation and evidence traces
of pathology agents~\cite{LimChi_PathTrace_MICCAISAT2026}.
Image provenance analysis reconstructs source relationships
and transformation histories among related images
~\cite{moreira2018image}.
CMPV instead verifies whether a magnified observation
originates from its claimed spatial location using
independent visual evidence.

\paragraph{Correspondence and uncertainty.}

Histopathology registration aligns tissue images across
sections or staining conditions
~\cite{borovec2020anhir,weitz2024acrobat,gatenbee2023virtual}.
CMPV instead evaluates a local observation's claimed
source, including cases where the source is absent
from the reference context.
LoFTR~\cite{sun2021loftr},
Efficient LoFTR~\cite{wang2024eloftr},
LightGlue~\cite{lindenberger2023lightglue},
and RoMa~\cite{edstedt2024roma}
establish geometric correspondences;
MatchAnything~\cite{he2025matchanything}
and MINIMA~\cite{ren2025minima}
extend matching across appearance changes.
DINOv2~\cite{oquab2023dinov2}
and UNI~\cite{chen2024uni}
provide frozen retrieval representations.
Relative match distinctiveness is used in
Lowe's ratio test~\cite{lowe2004sift}
and place-recognition uncertainty methods,
including SUE~\cite{zaffar2024sue}
and Statistical Uncertainty~\cite{miller2025doubt}.
ZoomTrace applies a related principle at the
candidate-region level, comparing the strongest
geometric match against spatially disjoint alternatives
to estimate source reliability.
Selective prediction~\cite{geifman2017selective}
provides an accuracy--coverage perspective for
comparing verifiers with different answering rates.

\section{Cross-Magnification Provenance Verification}
\label{sec:task}

\paragraph{Inputs and supervision.}
Let $C$ be a context image covering a slide region $B^c$ at physical resolution $m_c$, measured in micrometers per pixel. A query $Q$ is acquired at finer resolution $m_q<m_c$ and is claimed to originate from box $\ell$. The verifier receives $(C,Q,\ell,B^c,m_c,m_q)$. The benchmark additionally records the true source slide and box $B^*$, which are not available to the verifier. A correct claim identifies the true source; a provenance fault changes the source while retaining the claim. Image perturbations are recorded separately and do not change the underlying source label.

The context and its coordinate frame are trusted; actual query extraction coordinates are unavailable. Source recovery is defined only for sources inside the context.

\paragraph{Independent source estimation.}
A verifier estimates a source box $\widehat B$ and reliability $r$ from the images:
\begin{equation}
 (\widehat B,r)=f(C,Q;B^c,m_c,m_q).
 \label{eq:localise}
\end{equation}
The claimed box $\ell$ is excluded because its correctness is being evaluated. Estimating the source independently allows the verifier to identify spatial mismatches and recover an alternative source when one exists within the context.


\paragraph{Selective decisions.}
For an estimated box, define its discrepancy from the claim as
\begin{equation}
 d(\widehat B,\ell)=\frac{\|c(\widehat B)-c(\ell)\|_2}{w_q},
 \label{eq:residual}
\end{equation}
where $c(\cdot)$ denotes the box center and $w_q$ is the query footprint width in the slide coordinate frame. A reliability threshold $\tau$ and spatial tolerance $\delta$ produce the decision
\begin{equation}
 a=\begin{cases}
 \text{unverifiable}, & r<\tau,\\
 \text{verified}, & r\geq\tau,\ \widehat B\ne\varnothing,\ d\leq\delta,\\
 \text{mismatched}, & \text{otherwise}.
 \end{cases}
 \label{eq:decision}
\end{equation}
A missing estimate with $r\geq\tau$ yields a mismatch, while $r<\tau$ yields abstention. The ground truth remains binary; \emph{unverifiable} denotes the model's decision to abstain. For a mismatched query localized within the context, the estimated box also supports source correction.

\paragraph{Calibration.}
Each method fits $\delta$ to the 99th percentile of clean localization residuals and selects the lowest candidate $\tau$ that verifies at most 1\% of incorrect calibration claims. Thresholds remain fixed on test. Supplementary Section~\ref*{sec:protocol} gives threshold enumeration and edge cases.

\section{The PyraFault Benchmark}
\label{sec:benchmark}
PyraFault uses TCGA-LUAD, LUSC, LGG, BLCA, and UCEC~\cite{nci_tcga}, spanning four organ groups. Image-pyramid coordinates and slide identity determine the true source independently of the claim. Faults retain the requested location while substituting the source. Cases are separated into development, calibration, and test splits (Table~\ref{tab:data}), preventing case leakage. A $512^2$ query at $0.5\,\mu$m/px has a $128^2$ footprint in a $2048^2$ context at $2.0\,\mu$m/px.

\begin{table}[t]
\centering
\caption{The five-cohort PyraFault release. Splits are assigned by case. Examples represent query--claim pairs.}
\label{tab:data}
\small
\begin{tabular}{lr}
\toprule
Property & Count or setting \\
\midrule
TCGA cohorts / organ groups & 5 / 4 \\
Cases & 2,200 \\
Total examples & 3,057,732 \\
Development / calibration & 2,076,456 / 477,956 \\
Test & 503,320 \\
Cases: development / calibration / test & 1,540 / 330 / 330 \\
Context image & $2048^2$ at $2.0\,\mu$m/px \\
Query image & $512^2$ at $0.5\,\mu$m/px \\
\bottomrule
\end{tabular}
\end{table}
\paragraph{Provenance faults.}
Three \emph{within-context} faults permit local source recovery: adjacent drift (0.25--6 query widths), random within-context substitution, and hard within-context substitution mined for appearance similarity using frozen features. Substitutions use nonoverlapping alternatives. Four \emph{absent-source} faults use ordinary or appearance-mined hard alternatives outside the context on the same slide, or on another slide. Together with clean pairs, these give eight conditions sampled equally in test, each with 62,915 examples; correct claims form 12.5\% of the test set. Supplementary Section~\ref*{sec:construction} details candidate eligibility, donor strata, and quality control.

\paragraph{Viewing perturbations.}
Pairs receive one of four frozen conditions: none, mild, moderate, or severe. Perturbed queries rotate by $\pm5^\circ$, $\pm10^\circ$, or $\pm15^\circ$, with scale changes within $\pm2\%$, $\pm4\%$, or $\pm6\%$. Parameters are replayed at extraction and hidden from the verifier. These transformations preserve provenance labels, separating geometric robustness from source correctness.

\section{ZoomTrace}
\label{sec:method}

\begin{figure*}[t]
    \centering
    \includegraphics[width=\textwidth]{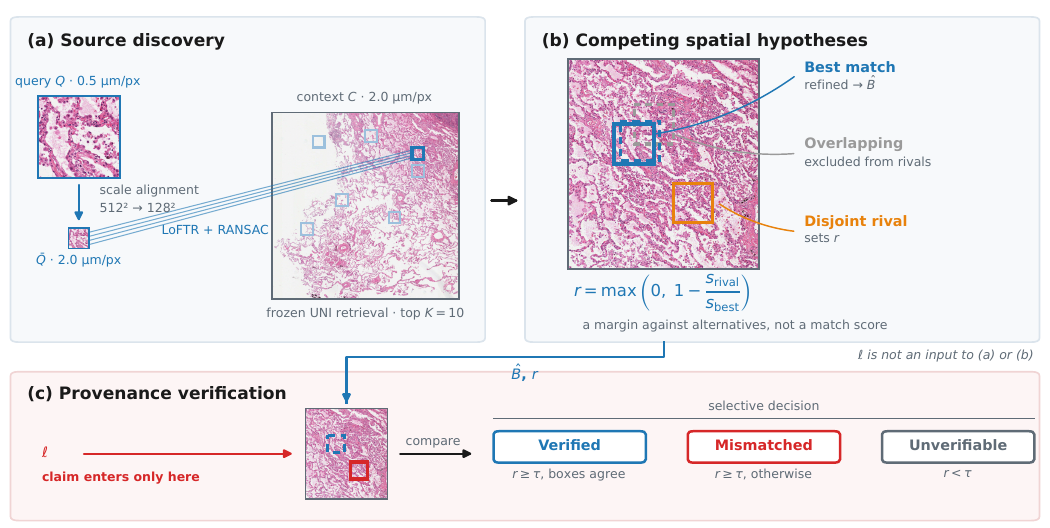}
    \caption{\textbf{Claim-independent source estimation and verification in ZoomTrace.}
    (a) The query is resampled to the context's physical resolution;
    frozen UNI features retrieve candidate windows, and LoFTR with
    RANSAC refines the source estimate $\widehat B$.
    (b) Reliability $r$ measures the winning match's margin over its
    strongest spatially disjoint retrieved competitor. Windows overlapping
    the winner are excluded from the rival set.
    (c) Only then is the claim $\ell$ introduced. Reliability threshold
    $\tau$ and spatial tolerance $\delta$ determine whether it is verified,
    mismatched, or unverifiable.
    The tissue comes from Figure~\ref{fig:motivation}; candidate windows
    are schematic placements on the retrieval grid.}
    \label{fig:method}
\end{figure*}

ZoomTrace estimates the source and reliability before consulting the claim (Figure~\ref{fig:method}), using frozen visual models.

\paragraph{Physical-scale alignment.}
A query with side length $H_q$ occupies $h=H_qm_q/m_c$ context pixels. We downsample it to this footprint before retrieval and matching. At the evaluated resolutions, $h=128$. A common physical scale exposes shared tissue structure.

\paragraph{Candidate retrieval.}
We place context windows $W_j$ of side $h$ on a grid of stride $h/2$ and represent each with a frozen UNI encoder~\cite{chen2024uni}. The overlapping grid reduces the displacement between an off-grid source and its nearest candidate, providing an initialization for geometric refinement. Cosine similarity between frozen UNI descriptors of the aligned query and each window ranks candidates.
The top ten windows form the candidate set $\mathcal K$. Context descriptors can be reused for subsequent queries from the same reference. For the evaluated context geometry, the bank contains 961 windows; only the ten retrieved candidates undergo geometric matching. This separates a reusable representation of the context from query-specific correspondence estimation. The retrieval grid covers the context independently of the claimed location.

\paragraph{Geometric refinement.}
For each retrieved window, LoFTR~\cite{sun2021loftr} establishes correspondences with the aligned query using frozen outdoor-pretrained weights. We fit a homography from window coordinates to query coordinates using RANSAC with a three-pixel threshold and score the candidate by its inlier fraction $s_j$. We select $j^*=\arg\max_{j\in\mathcal K}s_j$. If the winning window has origin $o_{j^*}$ and the fitted homography has translation entries $t_{j^*}$, the refined box has origin $o_{j^*}-t_{j^*}$ and retains the aligned query's side length $h$. The subtraction follows the window-to-query direction of the fitted transform. We round the refined origin to context pixels and map the box into slide coordinates using the known context origin and pixel-to-slide scale. A failed homography returns no source box.

\paragraph{Reliability from competing sources.}
A strong geometric match is less informative when another location fits almost equally well. We compare the winner against candidates with nonoverlapping windows:
\begin{equation}
 \mathcal R=\{j\in\mathcal K: W_j\cap W_{j^*}=\varnothing\}.
\end{equation}
For $s_{j^*}>0$ and a nonempty rival set, reliability is
\begin{equation}
 r=\max\!\left(0,1-\frac{\max_{j\in\mathcal R}s_j}{s_{j^*}}\right).
 \label{eq:margin}
\end{equation}
We assign zero reliability when the winning score is zero. With a positive score but no disjoint retrieved rival, the implementation assigns $r=1$. Reliability therefore measures separation within the retrieved candidate set. The rival set is defined against the winning retrieval window; its members can overlap one another.

Overlapping windows can represent the same source at shifted origins. Excluding them focuses the margin on competition between spatially distinct locations: a disjoint candidate with a similar score lowers reliability. Equation~\ref{eq:margin} reuses the geometric matching scores, requiring no additional matcher calls or learned confidence model. The claimed box enters only in Eq.~\ref{eq:decision}, which combines reliability with spatial agreement. A reliable estimate away from the claim supplies both a mismatch decision and a local source correction. Low reliability defers the decision even when the estimate overlaps the claim.

\section{Experiments}
\label{sec:experiments}
\paragraph{Setup and metrics.}
We evaluate nine frozen-model verifiers on all 503,320 test pairs: ZoomTrace; Efficient LoFTR~\cite{wang2024eloftr}, SuperPoint + LightGlue~\cite{detone2018superpoint,lindenberger2023lightglue}, MatchAnything--ELoFTR~\cite{he2025matchanything}, RoMa~\cite{edstedt2024roma}, and MINIMA--LightGlue~\cite{ren2025minima}; retrieval uncertainty methods SUE~\cite{zaffar2024sue} and Statistical Uncertainty~\cite{miller2025doubt}; and NCC~\cite{lewis1995template}. All share claim-independent estimation and calibration; a direct claim scorer is added in Section~\ref{sec:components}. For continuous discrimination,
\begin{equation}
 v=r\,\operatorname{IoU}(\widehat B,\ell),\qquad v=0\text{ if }\widehat B=\varnothing.
 \label{eq:vscore}
\end{equation}
AUPRC (average precision) and AUROC use all pairs, including abstentions, with correct claims positive. Decision coverage is the non-abstaining fraction; balanced accuracy (BA) is conditional on that subset. Localization uses median source IoU and the number of returned in-context estimates, independently of abstention. Protocols are in Supplement~\ref*{sec:protocol}.

\begin{table*}[!t]
\centering
\caption{CMPV on 503,320 test examples. Bold/underlined values denote best/second-best per metric; coverage and counts are unranked. AUPRC/AUROC use continuous scores; BA is conditional on answering. IoU is median source overlap over $N_{\mathrm{loc}}$ returned in-context estimates. R@.5 is the fraction of all 188,745 incorrect in-context claims localized at IoU$\geq0.5$, counting missing estimates as failures regardless of abstention. SUE and Statistical Uncertainty share retrieval. For the claim-conditioned baseline, R@.5 reflects overlap of the unchanged claimed box with the true source, not independent recovery. Dashes mark undefined BA and inapplicable independent localization. Full coverage for LightGlue/MINIMA includes missing-box mismatches (Supplement~\ref*{sec:sparsecoverage}).}
\label{tab:main}
\small
\setlength{\tabcolsep}{4pt}
\begin{tabular}{lrrrrrrr}
\toprule
Method & AUPRC $\uparrow$ & AUROC $\uparrow$ & Coverage & BA $\uparrow$ & IoU $\uparrow$ & $N_{\mathrm{loc}}$ & R@.5 $\uparrow$ \\
\midrule
\textbf{ZoomTrace (ours)} & \textbf{0.9686} & \underline{0.9872} & 0.3439 & 0.9847 & 0.7987 & 251,660 & \textbf{0.9726} \\
Efficient LoFTR & 0.7885 & 0.9026 & 0.3614 & \underline{0.9977} & 0.9616 & 208,334 & \underline{0.8275} \\
SUE & 0.6999 & 0.8672 & 0.0122 & 0.5127 & 0.5022 & 251,660 & 0.5034 \\
Statistical Uncertainty & 0.6487 & 0.8630 & 0.0109 & 0.5295 & 0.5022 & 251,660 & 0.5034 \\
NCC & 0.5940 & 0.7589 & 0.0099 & 0.5008 & 0.8399 & 251,660 & 0.5307 \\
MatchAnything--ELoFTR & 0.3625 & 0.6345 & 0.0882 & \textbf{0.9992} & 0.9615 & 71,174 & 0.2829 \\
SuperPoint + LightGlue & 0.2172 & 0.5507 & 1.0000 & 0.5533 & \underline{0.9825} & 27,438 & 0.1095 \\
MINIMA--LightGlue & 0.1439 & 0.5103 & 1.0000 & 0.5109 & \textbf{0.9848} & 5,567 & 0.0222 \\
RoMa v1 & 0.1269 & 0.5027 & 0.0101 & 0.4996 & 0.0000 & 153,475 & 0.0024 \\
\midrule
Claim-conditioned LoFTR & \underline{0.8838} & \textbf{0.9897} & 0.0000 & \textemdash & \textemdash & \textemdash & 0.0073 \\
\bottomrule
\end{tabular}
\end{table*}

\subsection{Benchmark Performance and Provenance Failures}
\label{sec:mainresults}
\paragraph{Discrimination and source recovery.}
ZoomTrace combines strong claim discrimination with broad source recovery (Table~\ref{tab:main}). It achieves AUPRC 0.969 and AUROC 0.987, versus 0.789 and 0.903 for Efficient LoFTR. Median source IoU is 0.799 over all 251,660 in-context examples; Efficient LoFTR reaches 0.962 on 208,334 estimates. Recovery recall instead counts missing estimates as failures on all 188,745 incorrect claims with in-context sources. At IoU$\geq0.5$, ZoomTrace recovers 97.26\%, versus 82.75\% for Efficient LoFTR (Supplement~\ref*{sec:recovery}). On 210,647 shared estimates, ZoomTrace also leads in discrimination: AUPRC 0.976 versus 0.926 and AUROC 0.987 versus 0.977 (Supplement~\ref*{sec:sharedsubset}).

\begin{figure}[t]
\centering
\includegraphics[width=\linewidth]{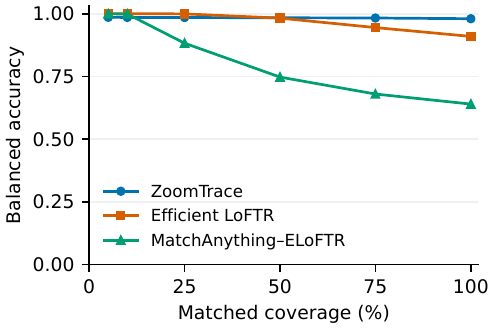}
\caption{Balanced accuracy on reliability-ranked subsets at six matched coverages. Each method retains its calibrated spatial tolerance. Points describe exact rank selections, not deployable threshold operating points.}
\label{fig:coverage}
\end{figure}

\paragraph{Selective decisions.}
\label{sec:matched}
At independently calibrated operating points, ZoomTrace has BA 0.985 at 34.39\% coverage; Efficient LoFTR has BA 0.998 at 36.14\%. Figure~\ref{fig:coverage} compares equal-sized subsets ranked by each method's reliability, retaining its spatial tolerance and stable input order within ties. Efficient LoFTR leads at 5--25\%; ZoomTrace leads at 50\%, 75\%, and 100\%, reaching full-coverage BA 0.980 versus 0.910. Rank-selected subsets can split ties and differ from calibrated operating points.

\begin{table}[t]
\centering
\caption{ZoomTrace by fault type. Each condition has 62,915 items. Both coverage and verified fraction use all items in the row.}
\label{tab:faults}
\small
\setlength{\tabcolsep}{4pt}
\begin{tabular}{lrr}
\toprule
Condition & Coverage & Verified \\
\midrule
Clean & 0.6731 & 0.6729 \\
Adjacent drift & 0.6775 & 0.0598 \\
Random within-context & 0.6751 & 0.0000 \\
Hard within-context & 0.6744 & 0.0000 \\
Out-of-context & 0.0118 & 0.0000 \\
Hard out-of-context & 0.0117 & 0.0030 \\
Out-of-slide & 0.0194 & 0.0000 \\
Hard out-of-slide & 0.0083 & 0.0003 \\
\bottomrule
\end{tabular}
\end{table}

\paragraph{Which faults are answered?}
Clean and within-context conditions have coverage near 67\% (Table~\ref{tab:faults}). Random and hard within-context substitutions have no reported false verifications. Adjacent drift has a verified fraction of 5.98\% of all drift items: small spatial displacements can fall within the nonzero tolerance $\delta\approx0.387$. Absent-source conditions have only 0.83--1.94\% coverage, so most yield abstention rather than an explicit mismatch. False-verification rates remain low: 0.30\% for hard out-of-context and 0.03\% for hard out-of-slide substitutions. Avoiding false verification differs from rejecting corrupted observations; only the latter enables middleware interception.

\subsection{Understanding ZoomTrace}
\label{sec:components}
\begin{table*}[t]
\centering
\caption{Component and reliability variants on all 503,320 test items, each calibrated separately. BA is conditional on answering (undefined at zero coverage). Calls count geometric matches per query; decoy adds 128.}
\label{tab:components}
\small
\begin{tabular}{lrrrrrr}
\toprule
Variant & AUPRC & AUROC & Cov. & BA & IoU & Calls \\
\midrule
Full ZoomTrace & \textbf{0.9686} & \underline{0.9872} & 0.3439 & \underline{0.9847} & \textbf{0.7987} & 10 \\
No scale alignment & 0.2012 & 0.5712 & 0.0105 & 0.5005 & 0.0000 & 10 \\
$K=1$ candidate & 0.9317 & 0.9703 & 0.0000 & -- & 0.7868 & 1 \\
No refinement & 0.9204 & 0.9825 & 0.1415 & 0.9543 & 0.5904 & 10 \\
Decoy statistic & \underline{0.9521} & \textbf{0.9888} & 0.4519 & \textbf{0.9858} & \underline{0.7986} & 138 \\
\bottomrule
\end{tabular}
\end{table*}

\paragraph{Scale and geometric localization.}
Scale alignment and geometric refinement both improve source localization and claim discrimination. Removing scale alignment reduces AUPRC to 0.201 and median IoU to zero (Table~\ref{tab:components}). Removing geometric refinement reduces these metrics to 0.920 and 0.590: returning a retrieved window restricts the estimate to the candidate grid.

\paragraph{Candidate diversity and residual tails.}
With $K=1$, retrieval remains active and median IoU is 0.787, but clean residual p99 increases from 0.387 to 5.464 query widths. The resulting tolerance and reliability calibration yield zero test coverage despite AUPRC 0.932. Fixing $\delta$ to the full model's value gives $K=1$ full-coverage BA 0.958 versus 0.980 for $K=10$. The residual tail explains the larger loss in calibrated coverage.

\begin{table*}[t]
\centering
\caption{Reliability on identical source estimates for 503,320 test items. Rules share $\delta=0.3756$ and fit separate $\tau$. This separate localization/calibration run differs from Table~\ref{tab:main} ($\delta=0.3872$), giving disjoint-margin coverage 0.4203 rather than 0.3439. (a) Aggregate discrimination and selective decisions; Range is maximum minus minimum coverage over four perturbation levels. (b) Coverage / conditional BA at those fixed thresholds. Top-two allows overlapping alternatives; Disjoint excludes them.}
\label{tab:relfixed}
\small
\textbf{(a) Aggregate results}\par\smallskip
\begin{tabular}{lrrrrr}
\toprule
Rule & AUPRC $\uparrow$ & AUROC $\uparrow$ & Cov. & BA $\uparrow$ & Range $\downarrow$ \\
\midrule
Best score & \textbf{0.9820} & \textbf{0.9916} & 0.4609 & \textbf{0.9861} & 0.1655 \\
Top-two & 0.5206 & 0.8274 & 0.4753 & 0.9702 & \underline{0.1371} \\
Disjoint & \underline{0.9686} & \underline{0.9872} & 0.4203 & \underline{0.9856} & \textbf{0.0228} \\
\bottomrule
\end{tabular}
\par\medskip
\textbf{(b) Coverage / conditional BA by geometric perturbation}\par\smallskip
\begin{tabular}{lrrrr}
\toprule
Rule & None & Mild & Moderate & Severe \\
\midrule
Winning score & 0.5232 / \textbf{0.9862} & 0.5141 / \textbf{0.9865} & 0.4485 / \underline{0.9861} & 0.3577 / \underline{0.9851} \\
Top-two & 0.4203 / 0.9581 & 0.4318 / 0.9666 & 0.4916 / 0.9729 & 0.5574 / 0.9737 \\
Disjoint margin & 0.4078 / \underline{0.9845} & 0.4211 / \underline{0.9855} & 0.4306 / \textbf{0.9862} & 0.4218 / \textbf{0.9863} \\
\bottomrule
\end{tabular}
\end{table*}

\paragraph{Reliability at fixed localization.}
Winning-score confidence gives stronger aggregate discrimination, while disjoint competition gives more uniform coverage across geometric perturbations. We compare three reliability rules on identical source estimates: the winning inlier fraction, its normalized margin over the strongest other candidate, and its margin over the strongest disjoint candidate (Table~\ref{tab:relfixed}). Each fits $\tau$ on the same calibration predictions and spatial tolerance; the caption distinguishes the runs.

Excluding overlapping candidates raises the margin's AUPRC from 0.521 to 0.969 (paired slide-bootstrap mean gain 0.448; 95\% CI $[0.438,0.459]$). Overlapping windows can represent the same source, so treating them as alternatives penalizes strong matches to that region. The winning score achieves higher AUPRC (0.982), AUROC (0.992), and conditional BA (0.986 at 46.09\% coverage), using the same ten matches.

With calibrated thresholds fixed across perturbation levels, the disjoint margin's coverage range is 0.023, versus 0.166 for the winning score (paired mean difference 0.142; 95\% CI $[0.134,0.151]$). Winning-score confidence answers more queries in the first three levels; under severe perturbation, the margin answers 42.18\% with BA 0.986, versus 35.77\% and 0.985. Table~\ref{tab:relfixed}(b) gives all coverage--accuracy pairs; Supplement~\ref*{sec:relfixed} gives paired intervals.

\paragraph{Sampled alternatives and matching cost.}
The decoy variant standardizes the winning score against 128 additional matches outside the refined source (Table~\ref{tab:components}). The margin has higher AUPRC and uses 10 rather than 138 matches; decoy scoring has higher AUROC and coverage. The runs produce different boxes on 3.45\% of items, so this compares complete variants; Table~\ref{tab:relfixed} isolates reliability on fixed estimates. Paired results and costs appear in Supplement~\ref*{sec:ablations}.

\paragraph{Claim independence.}
An implementation test holds context and query fixed and verifies that changing only the claim preserves the source box and reliability (Eq.~\ref{eq:localise}).

\paragraph{Direct verification at the claim.}
Claim-conditioned LoFTR aligns $Q$ and matches only the claimed crop using ZoomTrace's matcher. This single match yields AUPRC 0.884 and AUROC 0.990 (Table~\ref{tab:main}). ZoomTrace improves AUPRC by 0.085 (paired slide-bootstrap mean; 95\% CI $[0.079,0.090]$), while the direct baseline has higher AUROC. Adjacent drift is its most difficult fault, with AUROC 0.956 against clean; the other six faults achieve 0.988--0.999, including random and hard within-context substitutions. Its unchanged claimed boxes reach source IoU$\geq0.5$ on only 0.73\% of incorrect in-context claims, through residual overlap with the true source (Supplement~\ref*{sec:recovery}).

To apply the shared decision rule, this baseline sets $\widehat B=\ell$, making spatial discrepancy zero. This adapter can verify or abstain but cannot produce a mismatch or recover an alternative source. Moreover, 1.44\% of incorrect calibration claims attain the maximum score of 1, exceeding the 1\% false-verification budget. Calibration therefore selects $\tau=1+10^{-9}$, giving zero test coverage. For ZoomTrace, spatial agreement alone yields AUPRC 0.968, versus 0.240 for reliability alone and 0.969 for their product. Supplement~\ref*{sec:claimconditioned} reports implementation, costs, and fault-specific diagnostics.

\subsection{Robustness, VLMs, and Generalization}
\label{sec:strata}
\paragraph{Viewing perturbations.}
Selective decision accuracy remains high across geometric perturbations, although localization precision declines. ZoomTrace's coverage stays within 0.321--0.361 and conditional BA within 0.983--0.985, while median IoU falls from 0.990 to 0.607. Efficient LoFTR's coverage falls from 0.476 to 0.103. Refinement improves ZoomTrace's IoU at every level; without it, median IoU ranges from 0.605 to 0.569 (Supplement~\ref*{sec:perturbations}).

\begin{table}[t]
\centering
\caption{Claim-free localization on the same 2,000 items (11.8\% positive). Direct uses images; Grid adds an $8\times8$ reference grid.}
\label{tab:vlm_main}
\small
\begin{tabular}{llrr}
\toprule
Method & Input & AUPRC & AUROC \\
\midrule
ZoomTrace & Images & \textbf{0.9622} & \textbf{0.9857} \\
Efficient LoFTR & Images & \underline{0.7490} & \underline{0.8897} \\
Qwen3-VL-8B & Grid & 0.1188 & 0.5020 \\
Lingshu-32B & Direct & 0.1196 & 0.5041 \\
Claude Sonnet 5.5 & Grid & 0.1511 & 0.5546 \\
\bottomrule
\end{tabular}
\end{table}

\paragraph{Claim-free VLM localization.}
\label{sec:vlm}
Claim-free source localization remains difficult for the tested VLMs. Seven models receive the same 2,000 context--query pairs directly or with an $8\times8$ reference grid. Table~\ref{tab:vlm_main} includes Qwen3-VL~\cite{bai2025qwen3vl}, medical model Lingshu~\cite{xu2025lingshu}, and Sonnet~\cite{anthropic2026sonnet55}. The five open models have near-chance AUROC (0.497--0.511); API models show modest discrimination. ZoomTrace achieves AUPRC 0.962 on the same items. Full results and response diagnostics appear in Supplement~\ref*{sec:vlmdetails}.

\begin{table}[t]
\centering
\caption{CAMELYON16 (68,400 test items). Source-specific calibration is refitted for $K=50$. BA is conditional on answering; IoU uses returned in-context estimates.}
\label{tab:cam_main}
\small
\setlength{\tabcolsep}{4pt}
\begin{tabular}{lrrrrr}
\toprule
Method & AUPRC & AUROC & Cov. & BA & IoU \\
\midrule
ELoFTR & \underline{0.7849} & \textbf{0.8979} & 0.3154 & \textbf{0.9983} & \textbf{0.9610} \\
ZT, $K=10$ & 0.6761 & 0.8096 & 0.0129 & 0.5008 & 0.6536 \\
ZT, $K=50$ & \textbf{0.7918} & \underline{0.8835} & 0.0109 & \underline{0.5027} & \underline{0.7442} \\
\bottomrule
\end{tabular}
\end{table}

\paragraph{Calibration within TCGA.}
Localization tails affect coverage even within TCGA. Clean residual p99 is 4.595 query widths for LGG, versus 0.283--0.315 for other cohorts with similar central residuals. Fitting $\delta$ at p99 lets these failures widen spatial agreement, requiring a higher reliability threshold to control false verification (Supplement~\ref*{sec:detail-calibration}).

\paragraph{External-source evaluation.}
\label{sec:camelyon}\label{sec:calibration}\label{sec:tails}
CAMELYON16 uses the same geometry, fault taxonomy, and split protocol, with 68,400 test items. Efficient LoFTR outperforms ZoomTrace at the standard $K=10$ setting (Table~\ref{tab:cam_main}). ZoomTrace's AUPRC decreases to 0.676 and median IoU to 0.654; source-specific calibration yields coverage 0.013 and BA 0.501. The fraction of clean calibration queries localized more than one query width from the true source rises from 0.9\% on TCGA to 39.9\% on CAMELYON16. Two diagnostics separate retrieval from calibration effects.

First, retrieval misses true-source candidates. On 4,000 clean calibration queries per source, recall of windows with source IoU at least 0.5 is 81.7\% on TCGA and 35.9\% on CAMELYON16 at $K=10$. Widening to $K=50$ raises CAMELYON16 AUPRC to 0.792 and IoU to 0.744, with median localization time increasing from 150 to 535\,ms.

Second, clean residual p99 remains 12.618 query widths at $K=50$, versus 13.731 at $K=10$. This wide tolerance requires a high reliability threshold to control false verification, leaving coverage at 0.011. Reliability still correlates with residual ($-0.658$ Spearman); only 0.7\% of misplaced queries score above 0.5. Applying both TCGA thresholds to unchanged $K=10$ predictions gives coverage 0.175 and BA 0.983. These transferred thresholds are not fitted to the target-source constraint; the diagnostic separates calibration from retrieval (Supplement~\ref*{sec:recall}).

\subsection{Verification in the Visual Agent Tool Path}
\label{sec:agents}
\begin{table}[t]
\centering
\caption{Paired agent outcomes; ranks compare corrupted/protected paths within each agent, excluding the clean reference. E: fault exposure; S: silent failure (unflagged delivered corruption); A: verdict accuracy. E/S use all paired episodes; A uses episodes returning a verdict. $\tau_b=0.5009$ controls false verification; $\tau_w=0$ controls clean withholding.}
\label{tab:agents_main}
\small
\begin{tabular}{llrrr}
\toprule
Agent ($N$) & Tool path & E $\downarrow$ & S $\downarrow$ & A $\uparrow$ \\
\midrule
Claude (78) & Clean & 0.000 & 0.000 & 0.987 \\
 & Corrupted & 1.000 & 0.769 & 0.521 \\
 & Protected, $\tau_b$ & \underline{0.936} & \underline{0.667} & \underline{0.579} \\
 & Protected, $\tau_w$ & \textbf{0.538} & \textbf{0.423} & \textbf{0.756} \\
\midrule Codex (79) & Clean & 0.000 & 0.000 & 0.987 \\
 & Corrupted & 1.000 & 0.785 & \textbf{0.987} \\
 & Protected, $\tau_b$ & \underline{0.937} & \underline{0.734} & \textbf{0.987} \\
 & Protected, $\tau_w$ & \textbf{0.519} & \textbf{0.392} & \textbf{0.987} \\
\midrule DeepSeek (106) & Clean & 0.000 & 0.000 & 0.915 \\
 & Corrupted & 1.000 & 0.481 & \underline{0.755} \\
 & Protected, $\tau_b$ & \underline{0.972} & \underline{0.443} & 0.689 \\
 & Protected, $\tau_w$ & \textbf{0.453} & \textbf{0.179} & \textbf{0.829} \\
\bottomrule
\end{tabular}
\end{table}

\paragraph{Intervention and operating points.}
Verification reduces agent exposure to corruption, with verdict gains depending on the agent and operating point. ZoomTrace sits between crop extraction and observation delivery. It withholds mismatched observations and passes verified or unverifiable ones, so abstention still allows corrupted content to reach the agent. We compare the benchmark threshold $\tau_b\approx0.501$, controlling false verification, with $\tau_w=0$, selected under a 1\% budget for withholding clean views. The latter has calibration false-withhold rate 0.009. Both retain the same spatial tolerance; $\tau_w=0$ can withhold spatial disagreements that $\tau_b$ leaves unverifiable. Reference contexts use the benchmark resolution.

\paragraph{Exposure, silent failures, and verdicts.}
Paired evaluations connect Claude Code~\cite{anthropic_claudecode}, Codex~\cite{openai_codex}, and DeepSeek Harness~\cite{deepseek_harness} to the same visual tool server, using 78, 79, and 106 common valid episodes per arm. A silent failure occurs when corruption is delivered and the agent answers without flagging it or re-requesting the location. At $\tau_w$, exposure falls from 1.000 to 0.538, 0.519, and 0.453 (Table~\ref{tab:agents_main}); silent failures decrease from 0.769/0.785/0.481 to 0.423/0.392/0.179, respectively.

Claude Code's verdict accuracy improves from 0.521 to 0.756, with 19 corrected versus 3 newly incorrect verdicts (exact McNemar $p=8.6\times10^{-4}$). DeepSeek improves numerically from 0.755 to 0.829, with 20 corrected versus 12 newly incorrect ($p=0.22$); Codex remains at 0.987. Silent-failure reductions are significant for Claude and DeepSeek ($p=3.5\times10^{-6}$ and $2.1\times10^{-5}$). The benchmark threshold produces smaller exposure reductions and no uniform accuracy benefit: DeepSeek's accuracy is 0.689 at $\tau_b$, versus 0.755 without protection.

\paragraph{Practical execution.}
The PathAgent bridge measures 2.3\,s for reading a reference context, 5.7\,s for constructing its descriptor bank, and 0.5\,s for verification. The bank can be reused for subsequent queries within the same context.

\paragraph{PathAgent.}
At $\tau_b$, the PathAgent bridge~\cite{chen2025pathagent} withholds 12/31 adjacent-drift, 0/31 within-context, and 2/31 out-of-slide observations, no clean views withheld. Provenance answers remain unchanged across arms. Its precomputed navigation features limit the downstream effects observable from substituted pixels. See Supplement~\ref*{sec:agentdetails} for details.

\section{Conclusion}
\label{sec:conclusion}
CMPV and PyraFault make it possible to evaluate whether visual agents can trust the spatial origins of their magnified observations. ZoomTrace's independent localization supports claim discrimination and source correction; direct claim matching offers strong discrimination at lower cost. Reliability experiments reveal a trade-off between winning-score discrimination and more uniform coverage from disjoint competition. Tool-path verification reduces exposure, with verdict gains depending on the agent and threshold. Retrieval under source shift and calibration sensitivity to large residuals remain the main constraints on broader use.

\section*{LLM Usage Statement}
OpenAI Codex assisted with improving the clarity and presentation of the manuscript draft and refining the LaTeX layout. Claude Code assisted with implementing experimental protocols and writing the corresponding evaluation code. The VLMs and agent harnesses evaluated as experimental systems are identified in Section~\ref{sec:experiments} and the supplementary protocols.

{\small
\bibliographystyle{ieeenat_fullname}
\bibliography{main}
}
\end{document}